\documentclass{article}
\usepackage{times}
\usepackage{cite}
\usepackage{amsmath}
\onecolumn
\usepackage{graphicx}
\usepackage{latexsym}
\usepackage{subfigure}
\usepackage{pstricks}
\usepackage{setspace}
\date{}
\begin{document}
\title{Understanding Non-optical Remote-sensed Images: Needs, Challenges and Ways Forward\footnote{This is a white-paper submitted for the upcoming special issue in IEEE Signal Processing Magazine on Deep Learning for Visual  Understanding}}
\author{
Amit Kumar Mishra \\
Electrical Engineering Department \\
University of Cape Town, South Africa\\
Email: akmishra@ieee.org}

% make the title area

\maketitle
\begin{abstract}
Non-optical remote-sensed images are going to be used more often in managing disaster, crime  and precision agriculture. 
With more small satellites and unmanned air vehicles planning to carry radar and hyperspectral image sensors there is going to be an abundance of such data in the recent future. 
Understanding these data in real-time will be crucial in attaining some of the important sustainable development goals.  
Processing non-optical images is, in many ways, different from that of optical images. Most of the recent advances in the domain of image understanding has been using optical images. In this article we shall explain the needs for image understanding in non-optical domain and the typical challenges. Then we shall describe the existing approaches and how we can move from there to the desired goal of a reliable real-time image understanding system. \\
\emph{\textbf{remote sensing, big data, deep learning, cognitive architecture}}
\end{abstract}

\section{Introduction: Why do we need to understand non-optical images?}
Remote sensing (RS) using non-optical sensors like synthetic aperture radar (SAR) and radiometers has been becoming very popular as the cost of implementing such a system has been reducing over the past decade. 
This is evident from the number of new SAR sensor based small satellite projects started by various space agencies over the globe. 
The situation gets more interesting with the advent of low cost drones and the development in the domain of light airborne SAR sensor development. 
The driving motivation behind has been the extra information that a non-optical image delivers. 
This has been found very useful in precision agriculture. 
Secondly non-optical imaging sensors work under all weather condition and also in night. 
This makes such sensors particularly suitable for home-land security, and for the control of illegal trafficking. 
Both precision agriculture and control of illegal trafficking align closely to many of the sustainable development goals\cite{sus_dev}. 

Some of the major challenges of handling huge amount of RS non-optical images for extracting information are the lack of sufficient training data, and the highly dynamic nature of the scenes. 
For example  in many parts of the world locust infestation is a fairly unlikely event which occurs once in five to 10 years. %a decade. 
To predict such an event means we deal with hardly any training dataset. 
Secondly given the aggressive rate at which locusts infest the warning need to be delivered as quickly as possible. 
These are some of the challenges which can potentially be solved by proper image understanding schemes. 
Hence, the need for focussed image understanding research activities in the domain of non-optical RS images is an emerging and crucial field. 

There are some works in the open literature describing the application of emerging optical image processing tools in the domain of RS non-optical images. 
Some examples can be found in the references  \cite{tang_15, huang_97_gis, philip_92_back, han_15_opt, pal_13_kernel, pena_15_deep}.  
A recently published tutorial \cite{zhang_16_deep_tut} looks into many recently proposed deep learning algorithms and investigates their use in the domain of remote sensed (RS) image segmentation, classification and recognition. 
However this is a state-of-the-art review and does not discuss on why RS is unique and what more can be done to solve the typical challenges of RS image understanding.

\section{Plan of the final paper}
In addition to a detailed introduction describing the need and challenges of image understanding for remote-sensed non-optical images, the final submission will elucidate on the following themes. 

\subsection{Clarifying the Goal}
The major step towards solving any problem is to analyse the problem and the goal. 
This shall be explained following a system engineering approach\cite{nasa} where I shall start from ``user requirements'' and from there the functional requirements will be analysed and finally the desired specifications shall be presented. 

This shall elegantly make clear the bottlenecks in front of us and how we might try to overcome them. 

\subsection{Existing Approaches}
In this part some of the pertinent existing approaches shall be described. 
Only those algorithms which closely fit the ``goals'' shall be picked up. 
In describing them thrust will be put on the respective algorithms' strength and how the algorithms can be fine-tuned to fit our requirements. 

\subsection{How to get from where we are our goals}
In this last theme I shall describe some of the broad stream of approaches that can be taken to solve the problem of image understanding for non-optical RS images. 
This will be described under three headings. 
\subsubsection{Phenomenology and DeepNN}
A major difference between optical and non-optical images is the fact that optical images mosty represent the scene the way we ``see'' it. 
This helps a lot in the algorithm design phase of an image understanding algorithm. 
Especially for deep neural networks the deeper layers represent canonical and (mostly) invariant sub-features from the image. 
This is easier to arrive at than it is in non-optical images. 

In non-optical images arriving at canonical features requires a phenomenological study of the imaging system \cite{huynen, gab_1, gab_2}. 
This point shall be described in detail in this part. 

\subsubsection{Processing following the DIKW Pyramid}
Understanding, fundamentally, forms a part of the knowledge acquiring phase in the classic data-information-knowledge-wisdom (DIKW) pyramid. 
Putting the whole problem of image understanding in the DIKW pyramid makes the processing intuitive and easy to track. 
This will be detailed in this part. 

It can, however, be noted that this process might be equaly helpful for optical image understanding. 

\subsubsection{Cognitive architecture based global loop}
Human brain has been one of the major inspirations behind the development of much of the machine learning algorithms. 
Prefrontal cortex has been ascribed to as the source of human cognition and prefrontal cortex has a distinct layered architecture \cite{fuster_04, hebb_49}. 
This might be one of the reasons why deepNN performs so well. 
This kind of modelling has been well taken care of in cognitive architectures\cite{soar, lang_09} which deal with symbolic information processing.  

\begin{figure}[h]
\centering
\includegraphics[width=0.55\linewidth]{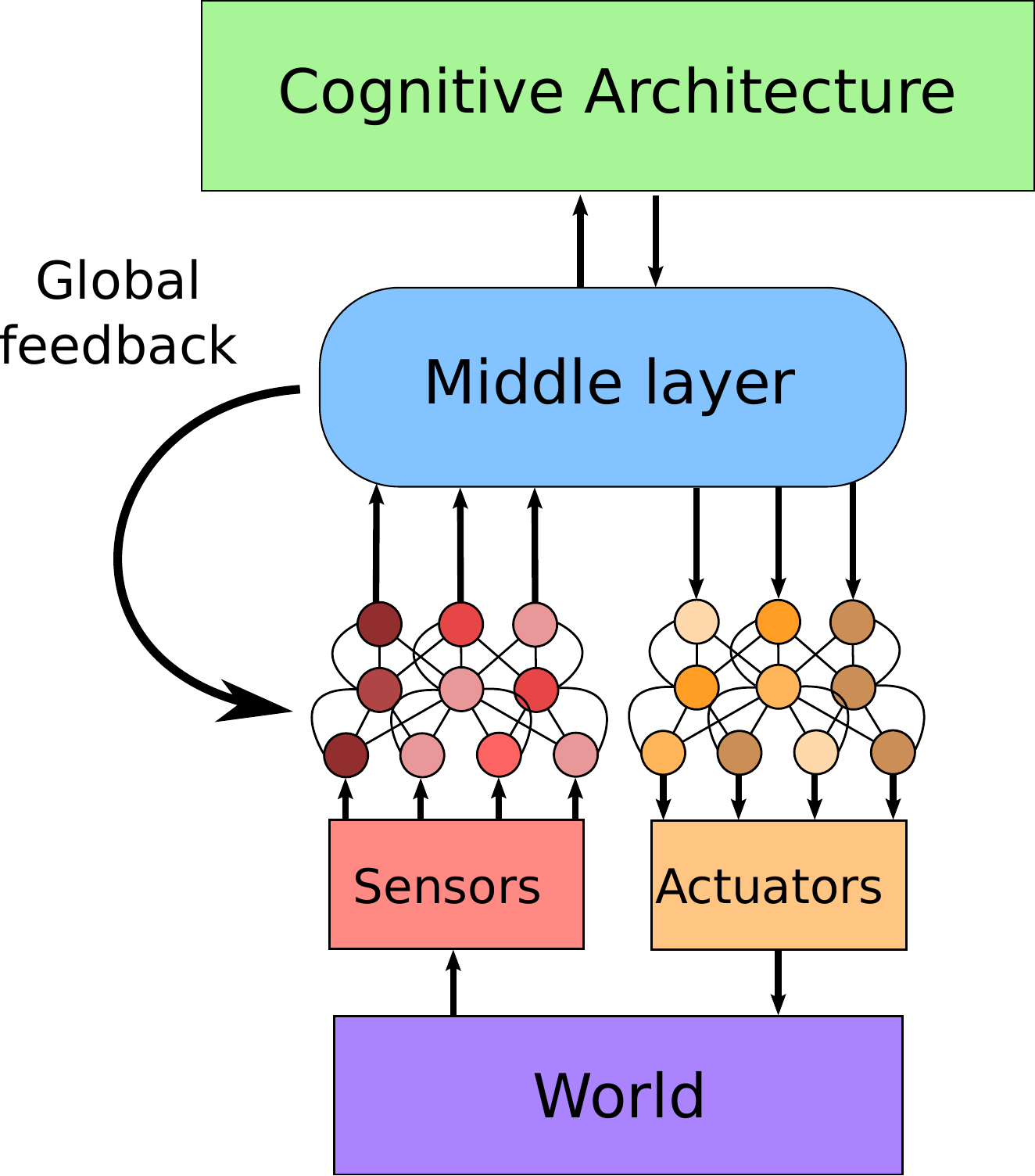}
\caption{A cognitive architecture can be used as a deliberative layer because of its high-level cognitive capabilities. The middle executive layer would control the flow of information between the deliberative and reactive layer. An ANN would make an appropriate reactive layer for perception and action. Further detail can be found in \cite{son_16}.}
\label{hyb_arch}
\end{figure}

We shall describe how a symbolic non-symbollic hybrid architecture\cite{son_16} inspired by prefrontal cortex can be used to process images for better understanding and for discovering knowledge from new images. 
The proposed structure can be seen in Figure~\ref{hyb_arch}.

\section{Conclusion}

\section*{Short Biography of the Author}
{\bf Dr. Amit Kumar Mishra} is an Associate Professor with the Radar Remote Sensing Group, University of Cape Town. 
He has more than 12 years of experience in the domain of radar system design and data analysis. 
In his work related to radar analysis he has worked extensively on synthetic aperture radar (SAR) image analysis and classification. 
He is a Senior Member of IEEE and has eight patents in which he is either the principal applicant or a co-applicant. 
One of his current interests is in the emerging domain of cognitive engineering and cognitive data processing. 

\bibliographystyle{IEEEtran}
\bibliography{ref}

%\begin{figure}[htbp]
%\includegraphics [scale=.4]{FIG3.eps}
%\caption{RMSE curves after Training and Validation}
%\end{figure}
%
%\begin{figure}[htbp]
%\includegraphics [scale=.4]{FIG4.eps}
%\caption{Membership functions before Training FIS }
%\end{figure}
%%\end{document}\\
%
%\begin{figure}[htbp]
%\includegraphics [scale=.4]{FIG5.eps}
%\caption{Membership functions after training FIS}
%\end{figure}

\end{document}